\documentclass[11pt]{article}
\usepackage[final]{acl}
\usepackage{times}
\usepackage{latexsym}
\usepackage[T1]{fontenc}
\usepackage[utf8]{inputenc}
\usepackage{microtype}
\usepackage{inconsolata}
\usepackage{amsmath,graphicx}
\hypersetup{hidelinks}
\graphicspath{{figs/}}

\title{Alphabet-Sequence Analogical Reasoning Across Large Language Models}
\author{Ryan Long, Wenrui Xu and Keshab K. Parhi \\ University of Minnesota, Twin Cities \\ \{long0585, xu000424, parhi\}@umn.edu\\}

\begin{document}
\maketitle

\begin{abstract}
Recent advances in analogical reasoning by large language models have raised the question of whether these capabilities extend across model families or remain specific to particular models. In this paper, we evaluate five frontier models on counterfactual letter-string analogies in a scrambled alphabet that covers six transformation tasks at two interval sizes, with 1,200 problems per model configuration. Under the evaluated settings, GPT-5.4 at high reasoning effort achieves the highest overall accuracy at both intervals (99.17\% and 97.33\%), followed by Kimi-K3, DeepSeek-V4-Pro, Qwen3.8-27B, and GLM-5.3. The overall ranking remains stable as \textit{interval size} increases, but performance declines vary substantially: GLM-5.3 falls from 58.67\% to 29.50\%, which is below the previously reported interval-2 human baseline of 51.5\%. Task-level results highlight sequence correction as a recurring difficulty across several models. Comparisons across GPT-5.4 reasoning-effort settings further demonstrate sensitivity to inference configuration, while longer inference latency across providers does not consistently correspond to higher accuracy. These findings establish a comparative baseline for analogical reasoning under the tested configurations and motivate further evaluation with matched inference budgets.
\end{abstract}

\section{Introduction}
Analogical reasoning is the capacity to recognize two superficially different situations that share an underlying relational structure, and to use that shared structure to transfer a rule, solution, or inference from a familiar domain (the ``base'') to an unfamiliar one (the ``target''). When a student is told that current in a circuit is like ``water flowing through a pipe'', the operative cognitive step is not memorization but structure transfer: identifying which relations stay fixed while surface details change freely. This capacity has occupied a privileged place in the study of intelligence since the earliest attempts to measure it formally. Spearman's original theory of intelligence rested on what he called the ``eduction of relations and correlates'', in effect, formal analogical inference, and Sternberg's later componential analysis of human abilities treated analogy problems as a primary window into the elementary information-processing steps underlying intelligence more broadly~\cite{spearman1923nature, sternberg1977intelligence}. Analogical reasoning is different from simple similarity judgment or lookup precisely because it is systematic and relational: the operation ``replace the last element with its successor'' should apply correctly whether it is instantiated over letters, numbers, or an entirely unfamiliar symbol system, which makes analogy a clean testbed for separating structural inference from surface-level pattern completion~\cite{Gentner1983}.

This property makes analogical reasoning a useful benchmark for large language models (LLMs), although distinguishing structural reasoning from memorized patterns remains challenging. A model that has only memorized surface statistics from its training corpus should fail when presented with a familiar transformation in an unfamiliar context. However, a model that has learned the underlying relational operation should not. In practice, distinguishing these two possibilities is difficult for any system trained on a large enough slice of the public internet to have encountered countless worked analogy problems close to their tested form. The goal of this paper is to explore whether the capacity for alphabet-sequence analogical reasoning in LLMs can be generalized across the current landscape of frontier language models (LMs) built by different laboratories with different training pipelines. In this paper, we evaluate five widely used LMs, including OpenAI's GPT-5.4~\cite{openai2026gpt54}, Alibaba's Qwen3.8-27B~\cite{qwen38}, Moonshot AI's Kimi-K3, DeepSeek's DeepSeek-V4-Pro~\cite{deepseekai2026deepseekv4}, and Zhipu AI's GLM-5.3~\cite{glm5team2026glm5vibecodingagentic}, on an identical counterfactual alphabet-sequence analogy benchmark to establish a comparative baseline for analogical reasoning across model families.

The paper is organized as follows. Section~\ref{sec: Background} reviews prior work on analogical reasoning and its evaluation in LMs. Section~\ref{sec: Method} describes the benchmark, inference configurations, and evaluation protocol. Section~\ref{sec: Results} presents the accuracy and inference latency results. Finally, Section~\ref{sec: Discussion} discusses the implications, limitations, and directions for future work.

\section{Background}~\label{sec: Background}
The question of whether strong performance on analogical reasoning tasks reflects genuine relational reasoning or reliance on learned statistical patterns remains an open question for LLMs. Back in 2023, Webb, Holyoak, and Lu~\cite{webb2023emergent} reported that GPT-3, tested zero-shot, matched or exceeded human performance across several analogy formats, including letter-string problems and a textual adaptation of Raven's Progressive Matrices, and argued this constituted evidence of an emergent, human-like analogical reasoning capacity rather than retrieval. This claim sits inside a broader debate over whether strong zero-shot performance reflects genuine abstraction or sophisticated statistical pattern matching that is difficult to tell apart from abstraction using in-distribution test items alone~\cite{Kumar2023}. Lewis and Mitchell~\cite{lewis2024using} tested this directly by constructing ``counterfactual'' variants of the letter-string task: the same relational structure, instantiated with alphabets reordered or otherwise perturbed away from the familiar $\{a, b, c,\dots, z\}$ sequence the models had seen innumerable times during training. GPT-3, GPT-3.5, and GPT-4 all showed sharply reduced accuracy on these counterfactual variants relative to the unperturbed originals, while human accuracy stayed comparably high across both conditions, a dissociation the authors read as evidence that the models' apparent analogical ability was not as robust or general as human analogical reasoning.

% Webb, Holyoak, and Lu~\cite{Webb2025} subsequently revisited this critique directly, retesting newer models on an expanded set of counterfactual letter-string problems posed in scrambled, often entirely fictional 26-symbol alphabets, and found that more capable models, particularly those able to supplement direct inference with code execution, recovered much of the gap Lewis and Mitchell had identified, which they took as renewed evidence for a genuine, if still imperfect, emergent analogical reasoning capacity. This is a scrambled-alphabet, counterfactual task design, instead of the original unperturbed letter-string benchmark, that the present study and its companion study adopt: every problem here is posed in a fictional, non-alphabetic-order 26-symbol sequence, specifically to keep the surface-familiarity confound identified by Lewis and Mitchell out of the comparison. Because this entire debate has so far been conducted almost exclusively within the GPT model family, however, it leaves open whether the counterfactual-robust analogical capacity Webb and colleagues report is a general property of large language models as a class, or one more closely tied to a single vendor's models than the ``emergent capability'' framing suggests. Resolving that question requires holding the task fixed while varying the model, the standard approach once a capability claim moves from a single demonstration to a claim about an entire model class in the broader LLM evaluation literature.

\citet{Webb2025} revisited this critique by evaluating language models on counterfactual letter-string analogies with permuted alphabets. They reported improved performance when models were equipped with code execution to support auxiliary operations such as counting, interpreting these findings as evidence that analogical reasoning can extend beyond familiar task representations. However, the role of tool support is important when comparing these results with those obtained under other inference configurations.
Our study adopts the scrambled-alphabet task design instead of the standard alphabetical ordering. All problems use a fixed permutation of the 26-letter alphabet, which reduces the usefulness of familiar alphabetic sequences while preserving the transformation rules under evaluation. The studies discussed above primarily focus on GPT-family models, which leaves the extent of comparable performance across other model families less well characterized. We therefore hold the task design and evaluation protocol fixed while comparing five LMs. This provides a controlled basis for describing cross-model performance differences under the tested configurations, without assuming that success on this benchmark establishes general analogical reasoning ability~\cite{Chang2024, Liang2023}.

\begin{table*}[!ht]
\centering
\small
\renewcommand{\arraystretch}{1.2}
\caption{Illustrative examples of the six transformation tasks at interval-1, assuming the fictional alphabet order $m \prec q \prec a \prec t \prec z$. Each example shows an individual transformation rather than a complete analogy problem.}
\resizebox{0.8\textwidth}{!}
{\begin{tabular}{p{0.21\textwidth}|p{0.55\textwidth}|p{0.24\textwidth}}
\hline
\textbf{Task} & \textbf{Description} & \textbf{Example} \\
\hline
\textit{Extend sequence} &
Append the next letter according to the specified alphabet order. &
\texttt{m q a} $\rightarrow$ \texttt{m q a t} \\
\hline
\textit{Successor} &
Replace the final letter with its successor in the specified alphabet. &
\texttt{m q a} $\rightarrow$ \texttt{m q t} \\
\hline
\textit{Predecessor} &
Replace the initial letter with its predecessor in the specified alphabet. &
\texttt{q a t} $\rightarrow$ \texttt{m a t} \\
\hline
\textit{Remove redundant letter} &
Remove repeated letters while preserving their sequence order. &
\texttt{m q q a t} $\rightarrow$ \texttt{m q a t} \\
\hline
\textit{Fix alphabetic sequence} &
Replace an incorrect letter to restore the intended alphabetic sequence. &
\texttt{m q z t} $\rightarrow$ \texttt{m q a t} \\
\hline
\textit{Sort} &
Reorder the letters according to the specified alphabet order. &
\texttt{t q m a} $\rightarrow$ \texttt{m q a t} \\
\hline
\end{tabular}}
\label{tab:transformation-tasks}
\end{table*}

\section{Methodology}~\label{sec: Method}
We evaluate five LMs on the scrambled-alphabet analogy benchmark of \citet{Webb2025}. Each problem presents an example transformation task and asks the model to apply the inferred rule to a new letter string. The benchmark consists of six transformation tasks: \textit{extend sequence}, \textit{successor}, \textit{predecessor}, \textit{remove redundant letter}, \textit{fix alphabetic sequence}, and \textit{sort}. Table~\ref{tab:transformation-tasks} provides illustrative examples of these transformations. All evaluations use the same permutation of the 26-letter alphabet and cover two \textit{interval sizes}. Interval-1 uses adjacent positions in the specified alphabet, and interval-2 uses positions separated by two steps. The interval manipulation changes the spacing relevant to the transformations while retaining the same six task categories.

% We evaluated five models on the same task design used by Webb et al. and by the GPT-5.4 follow-up study~\cite{Webb2025}: 100 trials of each of the six transformation tasks, at two difficulty tiers. Interval-1 problems require a one-step transformation in the fictional alphabet (e.g., replacing a letter with its immediate successor); interval-2 problems require a two-step transformation, increasing abstraction demands. Specifically, six transformation tasks are illustrated in Table.~\ref{tab:transformation-tasks}. Each model therefore completed 600 interval-1 problems and 600 interval-2 problems, all using the identical scrambled 26-letter alphabet, scored by exact match against the ground-truth answer with 95\% confidence intervals computed via the Wilson score method.

Each model configuration is evaluated on 100 problems per task at each interval size, which generates 600 problems per interval and 1,200 problems in total. Responses are scored by exact match against the ground-truth answers. We report task-level accuracy and overall accuracy for each interval size. The overall score gives equal weight to all six tasks due to the fact that each task contains the same number of problems. We compute 95\% Wilson confidence intervals using 100 observations for each task-level estimate and 600 observations for each overall estimate.
GPT-5.4 is evaluated at three reasoning-effort settings: \texttt{low}, \texttt{medium}, and \texttt{high}. Each of the other four models is evaluated under a single provider-default inference configuration. The primary cross-model comparisons use GPT-5.4's high-effort setting, while its low- and medium-effort results are reported separately to examine how performance varies with inference configuration. This analysis is motivated by prior work on inference-time scaling \citep{Snell2025}. However, inference budgets are not matched across providers, so the results should be interpreted as comparisons of the evaluated configurations rather than as performance under equal computational resources. 
% We include the human accuracy reported by \citet{Webb2025}, based on 99 participants at interval-1 and 97 participants at interval-2, as a contextual reference. 
In this paper, new human responses or matched human--model comparisons are not collected or conducted. 
Our primary analysis focuses on performance differences across the five LMs and their evaluated configurations.

% GPT-5.4 was evaluated at three reasoning-effort settings (``low,'' ``medium,'' ``high''); the four comparison models were queried through a single provider-default inference configuration, since fine-grained effort control is not uniformly exposed across vendor APIs. To keep the primary cross-model comparison on equal footing, we treat GPT-5.4's high-effort condition as its headline entry in the model-vs-model comparisons below, and report its low- and medium-effort results separately to quantify how much of its advantage is attributable to additional inference-time computation rather than to training-time knowledge~\cite{Snell2025}. Human baseline accuracy (99 participants, interval-1; 97 participants, interval-2), drawn directly from Webb et al.~\cite{Webb2025}, is reported for context only and is not treated as a primary axis of comparison, consistent with the benchmarking, instead of human-comparison, framing of this study.

\section{Results}~\label{sec: Results}
\begin{figure}[!ht]
\centering
\includegraphics[width=\linewidth]{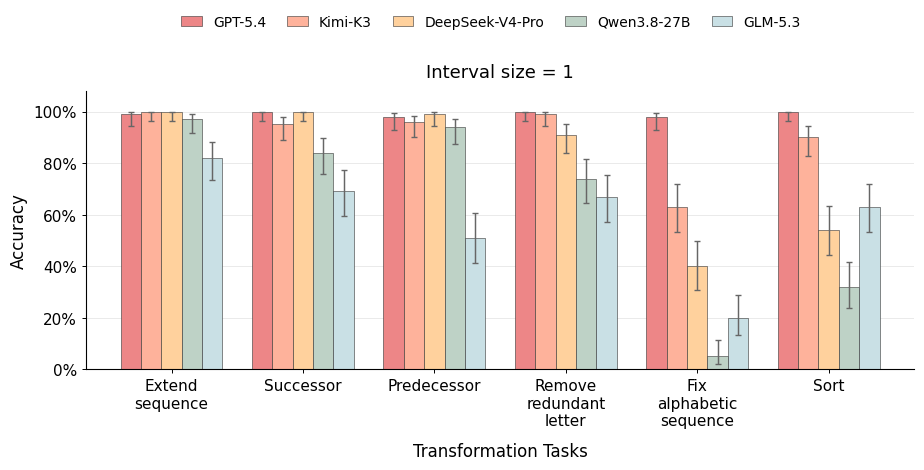}
\caption{Interval-1 accuracy across six transformation tasks for five LMs on the scrambled-alphabet benchmark of \citet{Webb2025}.}
\label{fig: interval-1}
\end{figure}

\begin{figure}[!ht]
\centering
\includegraphics[width=\linewidth]{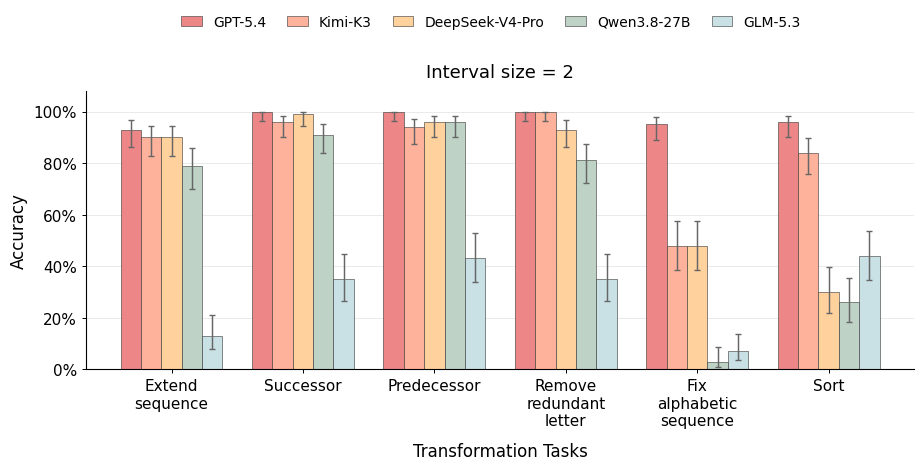}
\caption{Interval-2 accuracy across six transformation tasks for five LMs on the scrambled-alphabet benchmark of \citet{Webb2025}.}
\label{fig: interval-2}
\end{figure}

We examine model performance along three dimensions for all five LMs: overall accuracy at each interval size, variation across transformation tasks, and the relationship between accuracy and inference latency.  We evaluate each model configuration on 100 problems per transformation task, which amounts to 600 problems per interval size. The overall benchmark accuracy therefore weighs the six tasks equally. The main comparison uses GPT-5.4 at the high reasoning-effort setting, and the other four models at their provider-default configurations. The error bars in Figs.~\ref{fig: interval-1} and~\ref{fig: interval-2} denote 95\% Wilson confidence intervals for individual task accuracies.

% The overall ranking does not hold for every task. DeepSeek-V4-Pro trails Kimi-K3 by 9.83 percentage points overall, while it achieves higher accuracy on \textit{successor} (100\% vs.\ 95\%) and \textit{predecessor} (99\% vs.\ 96\%). In contrast, Kimi-K3 performs better on \textit{fix alphabetic sequence} (63\% vs.\ 40\%) and \textit{sort} (90\% vs.\ 54\%), while achieving at least 90\% accuracy on all tasks except sequence correction. Qwen3.8-27B indicates an even larger disparity, which achieves scores of 97\% on \textit{extend sequence} and 94\% on \textit{predecessor}, but only 5\% on \textit{fix alphabetic sequence}. Thus, similar performance on individual transformations can coexist with substantial differences in overall accuracy, which highlights the importance of reporting task-level results alongside aggregate scores.

\begin{figure*}[!ht]
\centering
\includegraphics[width=0.9\linewidth]{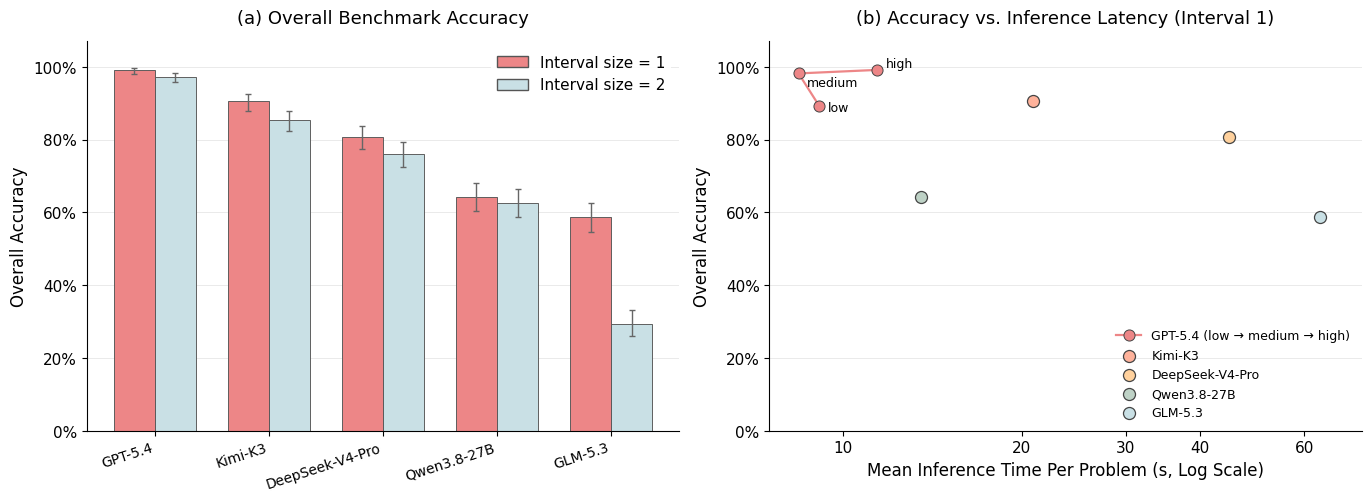}
\caption{Overall accuracy and inference latency across five LMs. (a) Accuracy at interval-1 and interval-2, with 95\% confidence intervals based on 600 problems per model per interval. GPT-5.4 uses the high reasoning-effort setting. (b) Interval-1 accuracy versus average inference time per problem, with latency shown in seconds on a logarithmic scale. Connected points indicate GPT-5.4's low, medium, and high reasoning-effort settings, and standard points represent the other four models under their provider-default configurations.}
\label{fig: fig-benchmark}
\end{figure*}

At interval-1, GPT-5.4 achieves the highest overall accuracy (99.17\%), followed by Kimi-K3 (90.50\%), DeepSeek-V4-Pro (80.67\%), Qwen3.8-27B (64.33\%), and GLM-5.3 (58.67\%). As shown in Fig.~\ref{fig: interval-1}, GPT-5.4 outperforms Kimi-K3 by 8.67\% and maintains accuracies between 98\% and 100\% across all six transformation tasks. The other models exhibit greater variation across tasks, with strong performance on some transformations offset by lower accuracy on others.
Fig.~\ref{fig: interval-2} presents the task-level accuracy at interval-2. The overall ranking remains unchanged: GPT-5.4 leads with 97.33\%, followed by Kimi-K3 (85.33\%), DeepSeek-V4-Pro (76.00\%), Qwen3.8-27B (62.67\%), and GLM-5.3 (29.50\%). Although overall accuracy decreases for all LMs, their sensitivity to the increased interval size differs substantially. GPT-5.4 and Qwen3.8-27B show small declines of 1.83\% and 1.67\%, respectively, while Kimi-K3 and DeepSeek-V4-Pro decline by 5.17\% and 4.67\%. GLM-5.3 exhibits the largest reduction, dropping by 29.17\% to approximately half its interval-1 accuracy.

% The \textit{fix alphabetic sequence} task is a recurring source of errors across LMs. At interval-1, it yields the lowest accuracy for Kimi-K3, DeepSeek-V4-Pro, Qwen3.8-27B, and GLM-5.3, and is tied with \textit{predecessor} as GPT-5.4's lowest-scoring task at 98\%. At interval-2, it remains the weakest task for Kimi-K3 (48\%), Qwen3.8-27B (3\%), and GLM-5.3 (7\%). DeepSeek-V4-Pro instead performs worst on \textit{sort} (30\%), and GPT-5.4 has its lowest accuracy on \textit{extend sequence} (93\%). These results identify sequence correction as a shared difficulty for most of these LMs, although the magnitude of the difficulty and the weakest task vary across models and interval sizes. One possible explanation is that sequence correction requires identifying a violation of the intended ordering and selecting an appropriate replacement instead of applying a local transformation at a predefined position. The observed accuracies alone might not be able to establish the reasoning processes responsible for these errors.

The \textit{fix alphabetic sequence} task is a recurring source of errors across tested models. At interval-1, it yields the lowest accuracy for four models and ties with \textit{predecessor} for GPT-5.4 (98\%). At interval-2, it remains the weakest task for Kimi-K3 (48\%), Qwen3.8-27B (3\%), and GLM-5.3 (7\%). DeepSeek-V4-Pro performs worst on \textit{sort} (30\%) and GPT-5.4 on \textit{extend sequence} (93\%). This pattern may reflect the challenge of identifying and repairing ordering violations, although accuracy alone does not establish the underlying cause.
% Fig.~\ref{fig: fig-benchmark}a summarizes overall accuracy across both interval tiers as a leaderboard. Fig.~\ref{fig: fig-benchmark}b plots accuracy against average inference time per problem on a log scale, using GPT-5.4's three effort levels to trace an accuracy/cost curve alongside the four single-configuration models. GPT-5.4 at low effort (89.33\% interval-1, 9.1 s/problem) already outperforms every non-GPT model's high-water mark except Kimi-K3, at a fraction of GLM-5.3's latency. GLM-5.3 is simultaneously the least accurate and the slowest model tested, requiring 198.7 s per problem on average at interval-2, roughly 5$\times$ DeepSeek-V4-Pro's latency and 20$\times$ Kimi-K3's, while achieving the lowest accuracy of any model at that tier. Latency and accuracy are therefore not correlated across providers in this benchmark; more inference time did not buy GLM-5.3 more correctness.
Fig.~\ref{fig: fig-benchmark}(a) compares overall accuracy across the two interval sizes, while Fig.~\ref{fig: fig-benchmark}(b) shows interval-1 accuracy against mean inference latency. At low reasoning effort, GPT-5.4 achieves 89.33\% accuracy at 9.1 seconds per problem, which outperforms all comparison models except Kimi-K3. Its high-effort configuration increases accuracy to 99.17\% at 11.4 seconds per problem. Across models, longer latency does not consistently correspond to higher accuracy: GLM-5.3 is both the slowest (63.8 seconds per problem) and the least accurate (58.67\%) at interval-1. These comparisons indicate that performance under the evaluated configurations, and wall-clock latency does not directly measure computational cost. In this case, we observe that GPT-5.4 exhibits lower inference latency at the medium-effort setting than at the low level. One possible explanation is that a moderate reasoning budget enables the model to identify an effective solution path more efficiently, while the low setting may follow less efficient reasoning trajectories despite using a smaller nominal reasoning budget. In contrast, the high setting may engage in more extensive deliberation, which results in increased inference latency. 

\section{Discussion \& Conclusion}~\label{sec: Discussion}
In this paper, our results highlight differences in analogical reasoning across model families. GPT-5.4 maintains high accuracy across tasks, while Kimi-K3 and DeepSeek-V4-Pro perform well on several transformations but show larger deficits in sequence correction or sorting. GLM-5.3 falls below the previously reported human baseline at interval-2 \citep{Webb2025}. Performance above this reference is therefore specific to the evaluated models and configurations, rather than a uniform property of LLMs.

% Second, the recurring difficulty of \textit{fix alphabetic sequence} and, for some models, \textit{sort}, across five independently trained systems suggests a shared frontier for the field rather than a gap specific to any one lab. Both transformations require tracking and reconciling information across the entire sequence rather than applying one substitution rule locally, while prior work demonstrates that additional inference-time computation can improve reasoning performance~\cite{Snell2025}. GPT-5.4's own accuracy/effort curve in Fig.~\ref{fig: fig-benchmark}b supports this reading: additional inference-time computation contributes to its performance, although this benchmark does not isolate the effects of training-time representations. GLM-5.3's results show that spending more wall-clock time per problem does not by itself close this gap, indicating that the relevant computation must be effectively targeted rather than merely extended.
We also identify \textit{fix sequence correction} and \textit{sort} tasks as recurring difficulties for several models. These tasks involve identifying or restoring ordering relations across a sequence, although the observed errors do not establish a common underlying mechanism. Motivated by prior work on inference-time scaling \citep{Snell2025}, our comparison of GPT-5.4 effort settings shows that inference configuration affects accuracy (Fig.~\ref{fig: fig-benchmark}(b)). However, longer latency across providers does not consistently correspond to better performance and should not be treated as a direct measure of computational effort. These findings provide a comparative account of analogical reasoning under the tested configurations and highlight the value of task-level evaluation alongside aggregate scores. Future work should examine whether the observed patterns persist across different alphabet permutations, composite transformations, and more closely matched inference budgets.

\newpage
\section*{Limitations}

Our evaluation uses a single scrambled alphabet and six individual transformation tasks. Consequently, the results do not establish robustness to other alphabet permutations, composite transformations, or broader forms of analogical reasoning. Strong performance on this benchmark should not be interpreted as evidence of general reasoning ability.
The inference configurations also limit cross-model comparisons. Only GPT-5.4 is evaluated at multiple reasoning-effort settings, while the other models use provider-default configurations. Computational budgets are not matched across providers, so performance differences cannot be attributed solely to model capabilities. Moreover, wall-clock latency can reflect serving infrastructure and generation behavior, making it an imperfect proxy for inference compute.
Finally, the human baseline is collected from a prior study rather than a newly collected, matched participant sample. We therefore only use it as a contextual reference. Our accuracy-based analysis also does not identify the internal reasoning processes responsible for correct responses or errors.

\newpage
% Bibliography style is supplied by acl.sty (acl_natbib).
\bibliography{paper_refs}

\end{document}